\documentclass[journal]{IEEEtran}
\usepackage{amsmath,epsfig,setspace,cite,graphicx}
\usepackage{epstopdf}
\usepackage{subfigure}
\usepackage{array}
\usepackage{multirow}
\usepackage{makecell}
\begin{document}
\title{Self-Commmittee Approach for Image Restoration Problems using 
Convolutional Neural Network}

\author{Byeongyong Ahn,
	and~Nam~Ik~Cho,~\IEEEmembership{Senior~Member,~IEEE}
\thanks{B. Ahn and N. I. Cho are with the Dept. of Electrical and Computer Engineering, Seoul National University,
	1, Gwanak-ro, Gwanak-Gu, Seoul 151-742, Korea and also
	affiliated with INMC
	(e-mail: clannad@ispl.snu.ac.kr; nicho@snu.ac.kr).}}

\markboth{Journal of \LaTeX\ Class Files,~Vol.~14, No.~8, August~2015}%
{Shell \MakeLowercase{\textit{et al.}}: Bare Demo of IEEEtran.cls for IEEE Journals}

\maketitle

\begin{abstract}
There have been many discriminative learning methods 
using convolutional neural networks (CNN) for several
image restoration problems, which learn the
mapping function from a degraded input to the clean output.
In this letter, we propose a self-committee method that can
find enhanced restoration results from the multiple trial of
a trained CNN with different but related inputs.
Specifically, it is noted that the CNN sometimes finds
different mapping functions when the input is transformed
by a reversible transform and thus produces different
but related outputs with the original. Hence averaging
the outputs for several different transformed inputs
can enhance the results as evidenced by the network
committee methods. Unlike the conventional committee approaches
that require several networks, the proposed method
needs only a single network.
Experimental results show that adding an additional transform as a committee
always brings additional gain on image denoising and single image
supre-resolution problems.
\end{abstract}


\IEEEpeerreviewmaketitle

\section{Introduction}
  
Image restoration problems are to estimate high-quality images from 
low-resolution or degraded ones, which are mostly ill-posed problems.
Hence conventional image restoration methods exploited
various kinds of image priors such as
gradient model \cite{rudin1992nonlinear,osher2005iterative,weiss2007makes}, 
wavelet model \cite{chang2000adaptive,remenyi2014image}, 
Markov random field (MRF) \cite{roth2005fields,lan2006efficient,li2009markov},
sparse representation \cite{elad2006image,mairal2009non,dong2013nonlocally} 
and nonlocal self similarity (NSS) prior \cite{buades2005non,dabov2007image,gu2014weighted}. 
Although these algorithms have shown promising results, 
they suffer from some drawbacks. First, some of the models are heuristically 
designed and they involve parameters that needs to be tuned by a user. 
Therefore the performance may often depend on the characteristics of
input image and parameters. Moreover, the methods find the optimal solution 
by solving complex optimization problems that are
mostly computationally expensive and also difficult to be parallelized.
 
In recent years, learning based methods that can overcome 
the above stated problems have been developed. For example,
Schmidt et al. \cite{schmidt2014shrinkage} proposed a cascade of shrinkage 
fields (CSF). The algorithm unifies the random field-based model and quadratic 
optimization into a learning framework.  Chen et al. \cite{chen2016trainable} 
proposed a trainable nonlinear reaction diffusion (TNRD) model. 
This method learns the parameters for a diffusion model by the gradient descent 
procedure. In addition, with the rapid progress of graphic processing units 
(GPU) programming and parallel processing, deep learning based 
image restoration methods have also attracted great attentions. 
Burger et al. \cite{burger2012image} proposed a multilayer perceptron 
(MLP) based denoising algorithm, which achives the competitive performance 
with prior model based methods. Dong et al. \cite{dong2016image} presented a convolutional 
neural network (CNN) based image super-resolution method, which
is shown to outperform the prior based methods. Kim et al. \cite{kim20162} 
proposed a skip-connection, which showed that learning the
residual image is more effective. Zhang et al. \cite{zhang2016beyond} 
developed a deep CNN for image denoising, which utilizes the residual 
learning and batch normalization \cite{ioffe2015batch}. 
This network shows the state-of-the art performance 
for many restoration problems including Gaussian image denoising, 
single image super-resolution (SISR) and JPEG image deblocking. 
Although the deep learning based methods are proven to be effective 
in many tasks, they also have some limitations. First, the training can be 
struck to a local minima and therefore the initial condition for the training 
affects the performance. Zhao et al.\cite{zhao2016loss} showed that local 
minima limits the network performance. Second, since the training aims to 
minimize only the pixel-based error, we do not know which prior or which
structure is well dealt with the newtork. In this respect,  it is shown that
combining some image priors \cite{burger2013learning} or
using multiple networks can improve the performance of 
restoration or classification problems \cite{ciresan2011convolutional}.
 
In this letter, we propose a committee approach that works at the
inference stage to enhance the performance of CNN based image 
restoration methods. The idea of ``network committees'' for a vision task was introduced in \cite{ciregan2012multi,ciresan2011convolutional}, and it was shown to
achive the best performance for  
MNIST digit classification problem \cite{lecun1998mnist}. 
The main idea of this method is to 
average the outputs of differently trained networks (called
network committees) to the same input,
which could alleviate the local minima problem and increase the performance. 
Our proposed method differs from the conventional committee approaches 
in constructing the committee members. 
Specifically, we use only a single network named \textit{base network},
and instead of preparing committees  
as the different networks, we define the committees as the outputs
of the network with differently transformed inputs.
Precisely, we note that the trained network
sometimes finds different feature map for the transformed input such 
as flipped or linearly transformed images and thus
produces different output (when inverse transformed). Thus 
we prepare several transforms, and the transformed inputs are passed
through the network and their outputs are used as committees. 
The outputs are averaged to be the final output. The proposed 
method is named self-committee network (SCN) in the sense that only
a single network is used. Experimental 
results show that the proposed method can improve the performance of the 
CNN based image restoration methods without additional training 
or fine-tuning. 
 
\begin{figure}
	\centering
	\includegraphics[width=8.5cm]{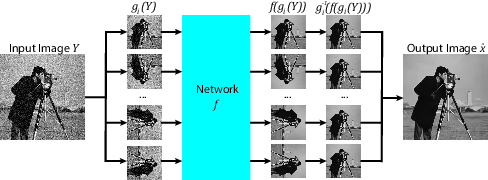}\\
	\caption{The structure of the proposed SCN framework.}
	\label{fig:Committee_Formulation}
\end{figure} 

\section{Proposed Algorithm}

The key ideas of our method are summarized as follows. Fist, some transformations 
are applied to an input image, which constructs a group of images for the given input. 
The group members are individually passed through the network and the outputs 
are inverse transformed to the original image space. Then the final output is 
estimated from the group of output images. An example of the proposed SCN 
framework is illustrated in Fig. \ref{fig:Committee_Formulation}.
In this letter, two kinds of image transformations are considered, which would
bring the output with almost the same performance but different characteristics.

\subsection{Flip and/or Rotation (FR)}
Training based image restoration algorithms \cite{schmidt2014shrinkage,chen2016trainable,burger2012image,
zhang2016beyond,dong2016image} aim to learn a mapping function 
$f(Y)\sim X$ for a degraded image $Y$ and its ground-truth $X$. In the view of 
human visual system (HVS), it is natural that the mapping function should
also work the same for the flipped and/or rotated image, i.e., it is desired that the 
FR image of the restored output must be the same as the restored output of the 
FR input : 
\begin{equation}
f(g(Y)) = g(f(Y))
\label{eq:Committee_invariance}
\end{equation}
where $g$ is the FR operation. Most prior based image restoration
methods satisfy this condition, because the FR 
operations do not affect the image prior such as gradient distribution or 
sparsity.

However, it does not hold for the CNN based image restoration methods.
 Although they  augment training data by FR operations
\cite{zhang2016beyond, chen2016trainable}, it does not force 
the trained convolution filters to be spatially symmetric, which is needed for
FR invariance. Therefore, they produce different results for the FR images
and thus it is worth to construct FR committees, where specific
operations are summarized in Table.~\ref{table:Committe_FR}.
In detail, we make \textit{member inputs} $\{g_{k}(Y)\}$ and their 
corresponding \textit{member outputs} $\{g_{k}^{-1}(f(g_{k}(Y)))\}$. 
Ciregan et al. \cite{ciregan2012multi} showed that averaging the outputs 
of the networks trained from the different initial states can improve the 
performance. Following the study, we also average the member outputs 
to get the final output
\begin{equation}
\hat{X}_{FR, I} = \frac{\sum_{k\in K}g_{k}^{-1}(f(g_{k}(Y)))}{\left| K \right| }
\label{eq:Committee_FR_Result}
\end{equation}
where $K$ is a subset of $\{1, 2, ...8\}$ and $\left| K \right|$ is the size of $K$.

\begin{table}
	\caption{8 FR operations employed to constitute the committee}
	\vspace{.0cm}
	\label{table:Committe_FR}
	\centering
	\setlength{\tabcolsep}{5pt}
	\begin{tabular}{ |c|c| }
		\hline
	  $k$ & Discription \\
		\hline
		\hline
		1 & Original \\
		\hline
		2 & FlipUD  \\
		\hline
		3 & Rotation $(90^\circ)$   \\
		\hline
		4 & Rotation $(90^\circ)$+FlipUD  \\	
		\hline
		5 & Rotation $(180^\circ)$\\
		\hline		
		6 & Rotation $(180^\circ)$+FlipUD  \\
		\hline
		7 & Rotation $(-90^\circ)$ \\
		\hline
		8 & Rotation $(-90^\circ)$+FlipUD  \\
		\hline		
	\end{tabular}
\end{table}

\subsection{Linearity}
Some image degradation models such as noise-free blurring or image downsampling 
are assumed as a linear model, $Y = XHV$ where $H$ is a blur kernel and 
$V$ is a resizing matrix. Therefore, it is natural that their corresponding 
restoration problems, i.e. deblurring or SISR, are also linear:

\begin{equation}
\label{eq:Committee_Invariance_SS}
f(\alpha Y + \beta) = \alpha f(Y) + \beta
\end{equation}
for any scalar $\alpha$ and $\beta$. 
However, the neural network assumes that the mapping function is 
non-linear and the network contains bias term in every neuron and 
non-linear activation functions such as rectified linear unit (ReLU). 
As a result,
(\ref{eq:Committee_Invariance_SS}) does not hold for neural network 
based algorithms, which will produce different outputs for 
the scaled and/or biased inputs (even when they are 
restored by removing the bias and rescaled). Hence we can prepare 
a committee for the member of inputs with several different  $\alpha$ and $\beta$,
i.e., we construct the output as
\begin{equation}
\label{eq:Committee_SS}
\hat{X}_{L} = \frac{\sum_{\alpha, \beta}\hat{x}_{\alpha, \beta}}{\sum_{\alpha, \beta}1}
\end{equation}
where
\begin{equation}
\label{eq:Committee_SS2}
\hat{x}_{\alpha, \beta} = \frac{f(\alpha Y + \beta) - \beta}{\alpha}.
\end{equation}
However, we cannot freely set the $\alpha$ and $\beta$
in the noisy environments $Y = XHV + N$ where $N$ is the noise,
because the scaling $\alpha$ changes the noise characteristics. 
Assuming that the noise distribution is zero mean and symmetric,
we can use just two committees such that  $\{(\alpha, \beta)\} = \{(1, 0), (-1, 1)\}$
for the noisy environment in order not to scale the noise component.
Specifically, we obtain the output as
\begin{equation}
\label{eq:Committee_Inv}
\hat{X}_{I} = \frac{f(Y)+(1-f(1-Y))}{2}
\end{equation}
which maintain the range of input pixel values, on which the network is 
trained and works best.

Since the linearity and FR invariance are independent property, 
they can also cooperate to make a larger committee as 

\begin{equation}
\hat{X}_{Full} = \frac{\sum_{\alpha, \beta}\sum_{k\in K}g_{k}^{-1}(f(g_{k}(\alpha Y+ \beta)))-\beta}{\sum_{\alpha, \beta}\alpha\left| K \right|}
\label{eq:Committee_FULL_Result}
\end{equation}

\section{Experimental Results}
We conducted experiments for two types of the image restoration: Image denoising and 
SISR. The performance is evaluated by the peak signal-to-noise ratio (PSNR) 
\cite{gonzalez2008digital} and improved PSNR (IPSNR) compared to the base network. 
We test 6 types of committees that are summarized in Table. \ref{table:Committe_Denoising}.

\begin{table}
	\caption{6 types of committee that are evaluated}
	\vspace{.0cm}
	\label{table:Committe_Denoising}
	\centering
	\setlength{\tabcolsep}{5pt}
	\begin{tabular}{ |c|c|c| }
		\hline
		\makecell{Committee \\ Name}& Discription & \makecell{$\sharp$ of \\Members} \\ 
		\hline
		\hline
		SCN-F & Original+Flip ($K=\{1, 2\}$) & 2\\ 
		\hline
		SCN-R & Original+Rotation ($K=\{1, 3, 5, 7\}$) & 4 \\
		\hline
		SCN-FR & Original+FR ($K=\{1\sim8\}$) & 8 \\
		\hline
		SCN-I & Original+Inversion & 2 \\
		\hline
		SCN-Full & Original+FR+Inversion & 16 \\
		\hline		
		SCN-L & Original+Linear (for SR only) & 3 \\
		\hline		
	\end{tabular}
\end{table}

\subsection{Experiments on image denoising network}
\begin{figure}
	\centering
	\begin{tabular}[t]{cccc}
		\includegraphics[width=1.7cm]{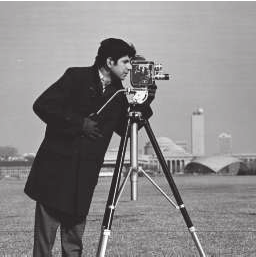}&
		\includegraphics[width=1.7cm]{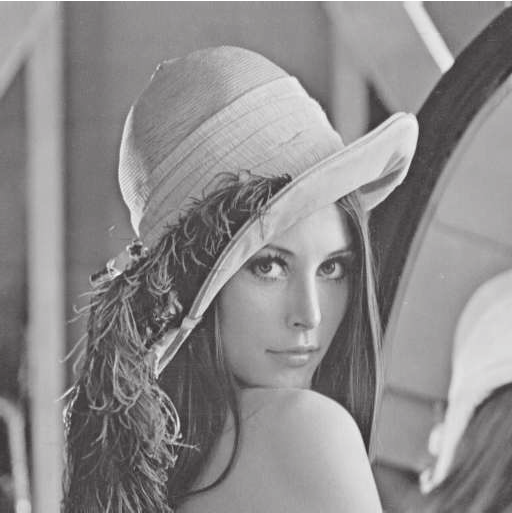}&
		\includegraphics[width=1.7cm]{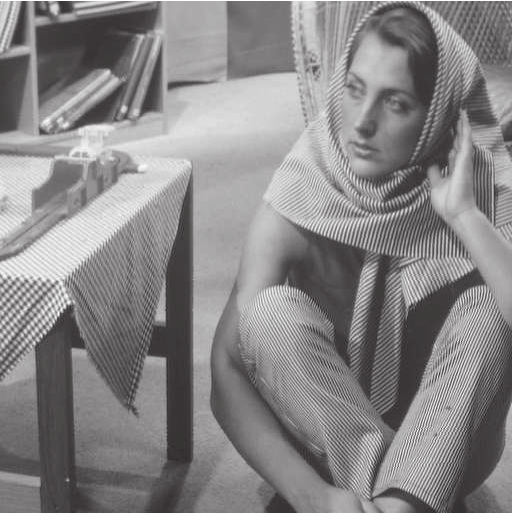}&
		\includegraphics[width=1.7cm]{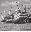}\\
		\includegraphics[width=1.7cm]{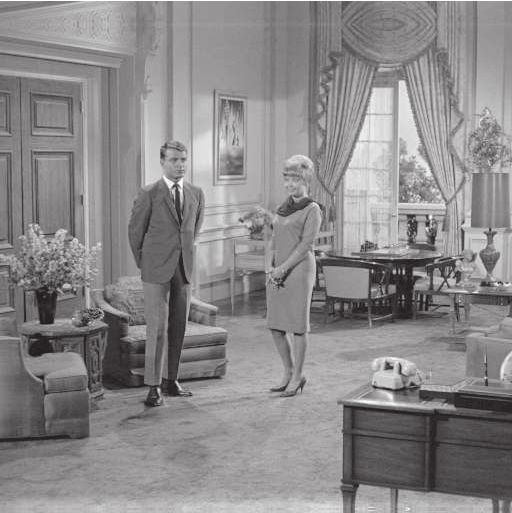}&
		\includegraphics[width=1.7cm]{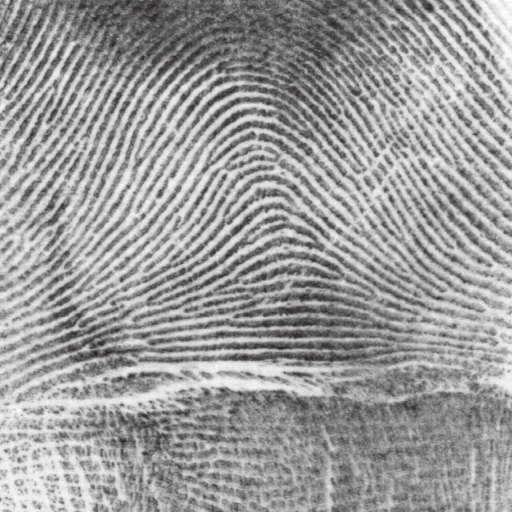}&
		\includegraphics[width=1.7cm]{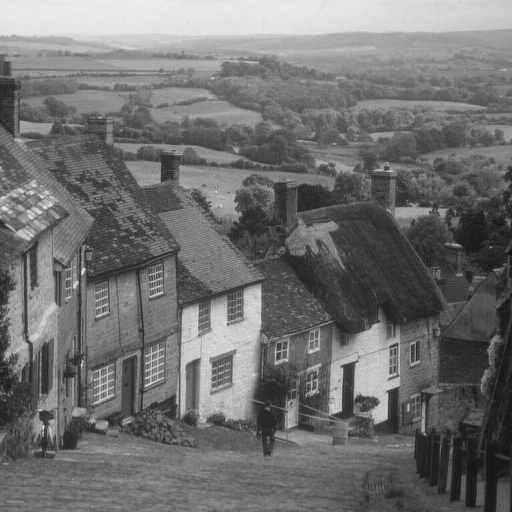}&
		\includegraphics[width=1.7cm]{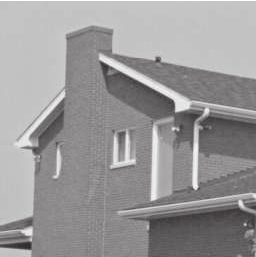}\\
		\includegraphics[width=1.7cm]{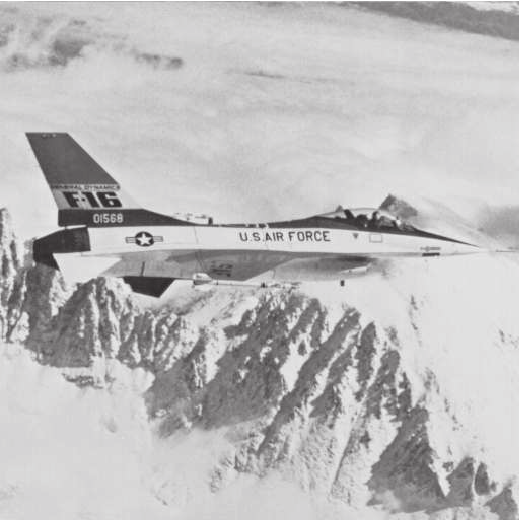}&
		\includegraphics[width=1.7cm]{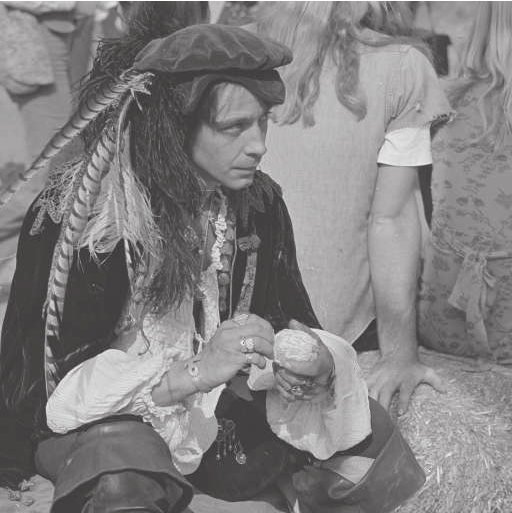}&
		\includegraphics[width=1.7cm]{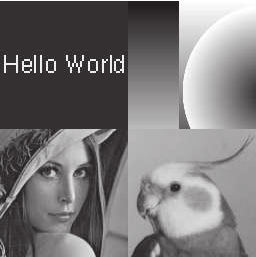}&
		\includegraphics[width=1.7cm]{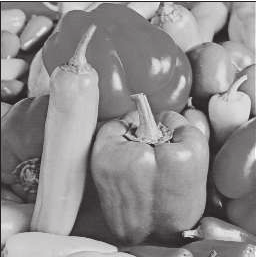}\\
	\end{tabular}
	\caption{The 12 test images used in the experiments}
	\label{fig:Committee_Testimg}
\end{figure}

\begin{figure}
	\centering
	\includegraphics[width=8.5cm]{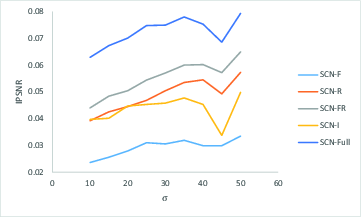}\\
	\caption{Average IPSNR curves for various SCN structures}
	\label{fig:Committee_IPSNR}
\end{figure}

For Gaussian image denoising, we use DnCNN \cite{zhang2016beyond} as a base network 
because of its promising performance and short run-time on GPU.  The test set is shown in 
Fig.~\ref{fig:Committee_Testimg}, which is consisted of 12 images that are widely used 
for the test of image denoising. Fig.~\ref{fig:Committee_IPSNR} summarizes the average 
IPSNR for various noise levels and Table. \ref{table:Committee_Result} shows the PSNR 
results on overall test images with $\sigma = 30$.

\begin{table*}[!]
	\caption{Individual PSNR results for gaussian denoising.}
	\vspace{.0cm}
	\label{table:Committee_Result}
	\centering
	\begin{tabular}{ |c|c|c|c|c|c|c|}
		\hline
		Method & DnCNN & SCN-F & SCN-R & SCN-FR & SCN-I & SCN-Full \\
		\hline
		\multicolumn{7}{c}{ $\sigma=30$}   \\
		\hline
		Cameraman & 29.24 & 29.26 & 29.28 & 29.28 & 29.28 & 29.30 \\
		\hline
		Lena & 31.62 & 31.66  & 31.67 & 31.68 & 31.66 & 31.69\\
		\hline
		Barbara & 28.84 & 28.89 & 28.93 & 28.94 & 28.91 & 28.96  \\
		\hline
		Boat & 29.36 & 29.38 & 29.40 & 29.40 & 29.38 & 29.41 \\		
		\hline
		Couple & 29.20 & 29.22 & 29.24 & 29.25 & 29.23 & 39.25 \\
		\hline		
		Fingerprint & 26.61 & 26.64 & 26.66 & 26.67 & 26.71 & 26.73 \\
		\hline
		Hill & 29.24 & 29.26 & 29.26 & 29.27 & 29.26 & 29.27\\
		\hline
		House & 32.38 & 32.43 & 32.43 & 32.44 & 32.42 & 33.45 \\
		\hline		
		Jetplane & 31.12 & 31.15 & 31.17 & 31.17 & 31.18 & 31.19 \\
		\hline
		Man & 29.23 & 29.25 & 29.26 & 29.27 & 29.24 & 29.26 \\
		\hline
		Montage & 31.82 & 31.89 & 31.93 & 31.95 & 31.87 & 31.94\\
		\hline	
		Peppers & 29.86 & 29.89 & 29.91 & 29.91 & 29.95 & 29.98 \\
		\hline
		\hline
		Average & 29.87 & 29.91 & 29.93 & 29.94 & 29.92 & 29.95 \\
		\hline
	\end{tabular}
\end{table*}

The results suggest the followings

\begin{itemize}
\item The employment of additional committee always improves the performance. 
\item The information of an image is severely distorted in a high noise level. 
Therefore, only a single network is hard to be optimal and adding the committees
is more beneficial at higher noise level. 
\end{itemize}

\begin{figure}
	\centering
		\begin{tabular}{cccc}
\multirow{2}{*}{Org.}
&
\multirow{2}{*}[6mm]{\includegraphics[width=2cm]{Dataset_Cameraman256.png}}
& {\includegraphics[width=2cm]{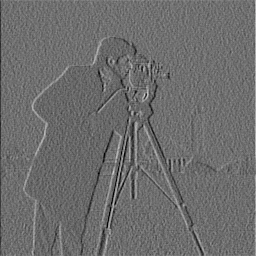}} &{\includegraphics[width=2cm]{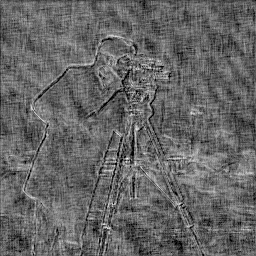}}  \\
&
& {\includegraphics[width=2cm]{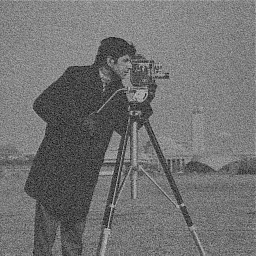}} &{\includegraphics[width=2cm]{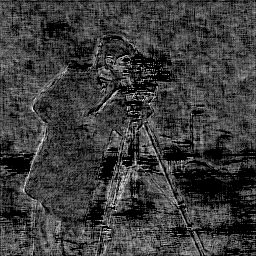}}  \\
\multirow{2}{*}{Inv.}
&
\multirow{2}{*}[6mm]{\includegraphics[width=2cm]{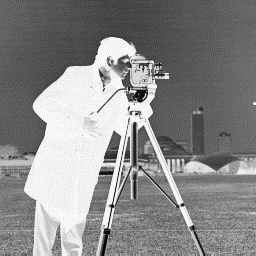}}
& {\includegraphics[width=2cm]{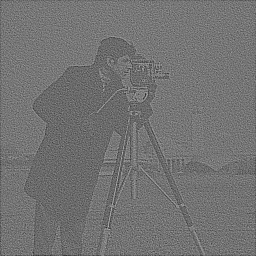}} &{\includegraphics[width=2cm]{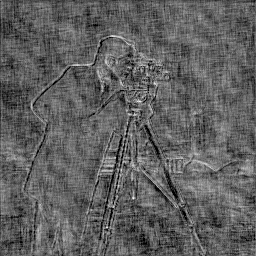}}  \\
&
& {\includegraphics[width=2cm]{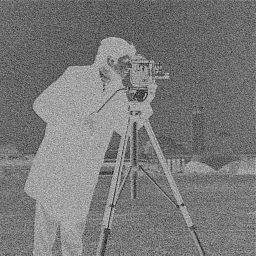}} &{\includegraphics[width=2cm]{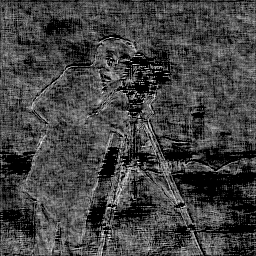}}  \\
& (a) & (b) & (c) \\ 
\end{tabular}

	\caption{The comparison of some features in the original image and inverted image.  (a) Input image, (b) features in the first layer, and (c) features in the 13-th layer.}
	\label{fig:Committee_Features}
\end{figure}

In order to analyze the improvement in view of the feature space, we extracted feature maps 
from an original image and its inverted one as illustrated in Fig.~\ref{fig:Committee_Features}. 
As shown in Fig. \ref{fig:Committee_Features} - (b), low-level feature maps of an inverted image 
are similar to the inversion of the original feature maps. However, the high-level features show 
somewhat different characteristics. The original feature map and inverted image feature map 
are similar in some cases (in the first and third row) but in other cases, they show weak 
correlation (in the second and fourth row). Moreover, the output of the inverted image 
would be re-inverted to the original image space and therefore, the two feature maps are 
distinct in the end. It implies that the function of a committee is expanding the feature maps 
and enables more accurate process, rather than just augmenting the input.

\begin{table*}[h]
	\caption{Average PSNR results for super-resolution}
	\vspace{.0cm}
	\label{table:Committee_SR}
	\centering
	\begin{tabular}{ |c|c|c|c|c|c|c|}
		\hline
		Dataset & \makecell{Upscaling \\ Factor} & SRCNN & SCN-FR & SCN-L &SCN-I & SCN-Full \\
		\hline
		\multirow{3}{*}{Set5} & 2 & 36.71 & 36.91 & 36.72 & 36.81 & 36.92 \\\cline{2-7}
		& 3 & 32.83 & 32.97 & 32.84 & 32.89 & 32.98 \\\cline{2-7}
		& 4 & 30.51 & 30.63 & 30.53 & 30.60 & 30.64 \\
		\hline
		\multirow{3}{*}{Set14} & 2 & 32.54 & 32.66 & 32.55 & 32.60 & 32.67 \\\cline{2-7}
		& 3 & 29.34 & 29.45 & 29.35 & 29.39 & 29.45\\\cline{2-7}
		& 4 & 27.52 & 27.58 & 27.53 & 27.57 & 27.59\\
		\hline
	\end{tabular}
\end{table*} 

\subsection{Experiments on a single image super-resolution network}
\begin{figure}
	\centering
	\begin{tabular}[t]{ccc}
		\includegraphics[width=2.5cm]{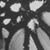}&
		\includegraphics[width=2.5cm]{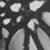}&
		\includegraphics[width=2.5cm]{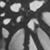}\\
		GT & SRCNN (32.87) & SCN-FR (33.21)\\ 
		\includegraphics[width=2.5cm]{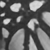}&
		\includegraphics[width=2.5cm]{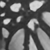}&
		\includegraphics[width=2.5cm]{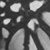}\\
		 SCN-L (32.97) &  SCN-I (33.05) & SCN-Full (33.23) \\ 
	\end{tabular}
	\caption{"Butterfly" image results with their PSNR}
	\label{fig:Committee_Test_SR}
\end{figure}

We also test the proposed SCN framework for a SISR. In order to show the 
robustness to the base network, we used SRCNN \cite{dong2016image} as a 
base network. We adopt two test datasets (Set 5 and Set 14) with three scaling 
factors (2, 3 and 4). Four committees as shown in 
Table.~\ref{table:Committe_Denoising} are tested: SCN-FR, SCN-I, SCN-L, and 
SCN-Full. For SCN-L, we set the parameters $\alpha$ and $\beta$ to
\begin{eqnarray}
&\alpha \in \{max(X) - min(X), 1,  \frac{1}{max(X) - min(X)}\}\\
&\beta = (1-\alpha)mean(X).
\label{eq:Committee_Linear}
\end{eqnarray}
By using these values, we can maintain the mean pixel value and prevent the 
pixel value saturation. Table~\ref{table:Committee_SR} lists the average PSNRs of 
different committees and Fig.~\ref{fig:Committee_Test_SR} presents an example. 
As shown in the results, the committee is beneficial for various the image restoration tasks 
and network formulations. Since the activation function (ReLU) keeps the linearity in a 
large range, scaling and shifting the input do not show notable difference. 
On the other hand, the inversion reverses the signs of the feature maps and 
thus draws out informations that are discarded from the original network. 
Hence the SCN-I generally yields higher PSNR than the SCN-L, which is
just a scaling based committee.

\section{Conclusion}
In this letter, we have presented a self-committee method to improve the performance of
CNN based image restoration algorithms.  Unlike the existing approaches
that use several differently trained networks as the committee members,
we use a single network and use the outputs of transformed inputs as the committee member.
The transfomed inputs induce different feature maps from the original,
and thus produces the outputs with different characteristics. Hence
averaging the outputs from differently transformed inputs could
enhance the restoration performances. Experiments show that 
the proposed method enhances the performance of state-of-the-art
image denoising and SISR networks.

\ifCLASSOPTIONcaptionsoff
  \newpage
\fi

\bibliographystyle{IEEEtran}
\bibliography{IEEEabrv,IEEEexample}

\begin{thebibliography}{10}
\providecommand{\url}[1]{#1}
\csname url@samestyle\endcsname
\providecommand{\newblock}{\relax}
\providecommand{\bibinfo}[2]{#2}
\providecommand{\BIBentrySTDinterwordspacing}{\spaceskip=0pt\relax}
\providecommand{\BIBentryALTinterwordstretchfactor}{4}
\providecommand{\BIBentryALTinterwordspacing}{\spaceskip=\fontdimen2\font plus
\BIBentryALTinterwordstretchfactor\fontdimen3\font minus
  \fontdimen4\font\relax}
\providecommand{\BIBforeignlanguage}[2]{{%
\expandafter\ifx\csname l@#1\endcsname\relax
\typeout{** WARNING: IEEEtran.bst: No hyphenation pattern has been}%
\typeout{** loaded for the language `#1'. Using the pattern for}%
\typeout{** the default language instead.}%
\else
\language=\csname l@#1\endcsname
\fi
#2}}
\providecommand{\BIBdecl}{\relax}
\BIBdecl

\bibitem{rudin1992nonlinear}
L.~I. Rudin, S.~Osher, and E.~Fatemi, ``Nonlinear total variation based noise
  removal algorithms,'' \emph{Physica D: Nonlinear Phenomena}, vol.~60, no.
  1-4, pp. 259--268, 1992.

\bibitem{osher2005iterative}
S.~Osher, M.~Burger, D.~Goldfarb, J.~Xu, and W.~Yin, ``An iterative
  regularization method for total variation-based image restoration,''
  \emph{Multiscale Modeling \& Simulation}, vol.~4, no.~2, pp. 460--489, 2005.

\bibitem{weiss2007makes}
Y.~Weiss and W.~T. Freeman, ``What makes a good model of natural images?'' in
  \emph{IEEE Conference on Computer Vision and Pattern Recognition(CVPR)},
  2007, pp. 1--8.

\bibitem{chang2000adaptive}
S.~G. Chang, B.~Yu, and M.~Vetterli, ``Adaptive wavelet thresholding for image
  denoising and compression,'' \emph{IEEE Transactions on Image Processing},
  vol.~9, no.~9, pp. 1532--1546, 2000.

\bibitem{remenyi2014image}
N.~Remenyi, O.~Nicolis, G.~Nason, and B.~Vidakovic, ``Image denoising with 2d
  scale-mixing complex wavelet transforms,'' \emph{IEEE Transactions on Image
  Processing}, vol.~23, no.~12, pp. 5165--5174, 2014.

\bibitem{roth2005fields}
S.~Roth and M.~J. Black, ``Fields of experts: A framework for learning image
  priors,'' in \emph{IEEE Conference on Computer Vision and Pattern
  Recognition(CVPR)}, vol.~2, 2005, pp. 860--867.

\bibitem{lan2006efficient}
X.~Lan, S.~Roth, D.~Huttenlocher, and M.~J. Black, ``Efficient belief
  propagation with learned higher-order markov random fields,'' in
  \emph{European Conference on Computer Vision(ECCV)}, 2006, pp. 269--282.

\bibitem{li2009markov}
S.~Z. Li, \emph{Markov random field modeling in image analysis}.\hskip 1em plus
  0.5em minus 0.4em\relax Springer Science \& Business Media, 2009.

\bibitem{elad2006image}
M.~Elad and M.~Aharon, ``Image denoising via sparse and redundant
  representations over learned dictionaries,'' \emph{IEEE Transactions on Image
  Processing}, vol.~15, no.~12, pp. 3736--3745, 2006.

\bibitem{mairal2009non}
J.~Mairal, F.~Bach, J.~Ponce, G.~Sapiro, and A.~Zisserman, ``Non-local sparse
  models for image restoration,'' in \emph{IEEE International Conference on
  Computer Vision(ICCV)}, 2009, pp. 2272--2279.

\bibitem{dong2013nonlocally}
W.~Dong, L.~Zhang, G.~Shi, and X.~Li, ``Nonlocally centralized sparse
  representation for image restoration,'' \emph{IEEE Transactions on Image
  Processing}, vol.~22, no.~4, pp. 1620--1630, 2013.

\bibitem{buades2005non}
A.~Buades, B.~Coll, and J.-M. Morel, ``A non-local algorithm for image
  denoising,'' in \emph{IEEE Conference on Computer Vision and Pattern
  Recognition(CVPR)}, vol.~2, 2005, pp. 60--65.

\bibitem{dabov2007image}
K.~Dabov, A.~Foi, V.~Katkovnik, and K.~Egiazarian, ``Image denoising by sparse
  3-d transform-domain collaborative filtering,'' \emph{IEEE Transactions on
  image Processing}, vol.~16, no.~8, pp. 2080--2095, 2007.

\bibitem{gu2014weighted}
S.~Gu, L.~Zhang, W.~Zuo, and X.~Feng, ``Weighted nuclear norm minimization with
  application to image denoising,'' in \emph{IEEE Conference on Computer Vision
  and Pattern Recognition(CVPR)}, 2014, pp. 2862--2869.

\bibitem{schmidt2014shrinkage}
U.~Schmidt and S.~Roth, ``Shrinkage fields for effective image restoration,''
  in \emph{IEEE Conference on Computer Vision and Pattern Recognition(CVPR)},
  2014, pp. 2774--2781.

\bibitem{chen2016trainable}
Y.~Chen and T.~Pock, ``Trainable nonlinear reaction diffusion: A flexible
  framework for fast and effective image restoration,'' \emph{IEEE Transactions
  on Pattern Analysis and Machine Intelligence}, 2016.

\bibitem{burger2012image}
H.~C. Burger, C.~J. Schuler, and S.~Harmeling, ``Image denoising: Can plain
  neural networks compete with bm3d?'' in \emph{IEEE Conference on Computer
  Vision and Pattern Recognition(CVPR)}, 2012, pp. 2392--2399.

\bibitem{dong2016image}
C.~Dong, C.~C. Loy, K.~He, and X.~Tang, ``Image super-resolution using deep
  convolutional networks,'' \emph{IEEE Transactions on Pattern Analysis and
  Machine Intelligence}, vol.~38, no.~2, pp. 295--307, 2016.

\bibitem{kim20162}
J.~Kim, J.~K. Lee, and K.~M. Lee, ``Accurate image super-resolution using very
  deep convolutional networks,'' in \emph{Proc. of Computer Vision and Pattern
  Recognition (CVPR)}, 2016, pp. 1646--1654.

\bibitem{zhang2016beyond}
K.~Zhang, W.~Zuo, Y.~Chen, D.~Meng, and L.~Zhang, ``Beyond a gaussian denoiser:
  Residual learning of deep cnn for image denoising,'' \emph{IEEE Transactions
  on image Processing}, 2017.

\bibitem{ioffe2015batch}
S.~Ioffe and C.~Szegedy, ``Batch normalization: Accelerating deep network
  training by reducing internal covariate shift,'' in \emph{Proc. of
  International Conference on Machine Learning (ICML)}, 2015, pp. 448--456.

\bibitem{zhao2016loss}
H.~Zhao, O.~Gallo, I.~Frosio, and J.~Kautz, ``Loss functions for image
  restoration with neural networks,'' \emph{IEEE Transactions on Computational
  Imaging}, 2016.

\bibitem{burger2013learning}
H.~C. Burger, C.~Schuler, and S.~Harmeling, ``Learning how to combine internal
  and external denoising methods,'' in \emph{German Conference on Pattern
  Recognition}, 2013, pp. 121--130.

\bibitem{ciregan2012multi}
D.~Ciregan, U.~Meier, and J.~Schmidhuber, ``Multi-column deep neural networks
  for image classification,'' in \emph{IEEE Conference on Computer Vision and
  Pattern Recognition(CVPR)}, 2012, pp. 3642--3649.

\bibitem{ciresan2011convolutional}
D.~C. Ciresan, U.~Meier, L.~M. Gambardella, and J.~Schmidhuber, ``Convolutional
  neural network committees for handwritten character classification,'' in
  \emph{IEEE Conference on Document Analysis and Recognition (ICDAR)}, 2011,
  pp. 1135--1139.

\bibitem{lecun1998mnist}
Y.~LeCun, C.~Cortes, and C.~J. Burges, ``The mnist database of handwritten
  digits,'' 1998.

\bibitem{gonzalez2008digital}
R.~C. Gonzalez, ``Digital image processing,'' 2008.

\end{thebibliography}

\end{document}